%% file: neurips_2019.tex
\documentclass{article}

     \usepackage[preprint]{neurips_2019}

\usepackage[utf8]{inputenc} % allow utf-8 input
\usepackage[T1]{fontenc}    % use 8-bit T1 fonts
\usepackage{hyperref}       % hyperlinks
\usepackage{url}            % simple URL typesetting
\usepackage{booktabs}       % professional-quality tables
\usepackage{amsfonts}       % blackboard math symbols
\usepackage{amsmath}
\usepackage{graphicx}
\usepackage{nicefrac}       % compact symbols for 1/2, etc.
\usepackage{microtype}      % microtypography
\usepackage[linesnumbered,vlined,ruled]{algorithm2e}
\usepackage{bm}
\usepackage{multirow}
\usepackage{float}

\title{[Re] Learning to Learn By Self-Critique}

\author{%
  Isac Arnekvist\thanks{Denotes equal contribution.} \\
  Divison of Robotics, Perception and Learning\\
  KTH Royal Institute of Technology\\
  \texttt{isacar@kth.se} \\
  \And
  Dmytro Kalpakchi\footnotemark[1]\\
  Division of Speech, Music and Hearing\\
  KTH Royal Institute of Technology\\
  \texttt{dmytroka@kth.se} \\
}

\begin{document}

\maketitle

\begin{abstract}
  This work is a reproducibility study of the paper of \cite{antoniou2019learning}, published at NeurIPS 2019. Our results
  are in parts similar to the ones reported in the original paper, supporting the central claim of the paper that the proposed
  novel method, called Self-Critique and Adapt (SCA), improves the performance of MAML++.
  The conducted additional experiments on the Caltech-UCSD Birds 200 dataset confirm
  the superiority of SCA compared to MAML++. In addition, the reproduced paper suggests a novel
  high-end version of MAML++ for which we could not reproduce the same results. We hypothesize that this
  is due to the many implementation details that were omitted in the original paper.
\end{abstract}

\section{Introduction}
Humans are very good at learning new concepts by observing just a few examples of each one. Modern deep learning methods are also good at learning new concepts, but require much more examples to learn any concept, albeit a very simple one, e.g. distinguishing cats from dogs. In striving to bridge the gap between the established supervised learning paradigm and fast-learning humans, the paradigm of \emph{few-shot learning} has emerged recently. In this new setting, the training data for each concept (or task) consists only of a few samples, called \emph{shots} (hence the name). The aim of the few-shot learning is to learn a variety of tasks with a few shots each, instead of learning one task with many shots (as in classical supervised learning). 

A popular way to approach few-shot learning is to frame it as \emph{meta-learning or learning to learn}, i.e., a learning paradigm focused on acquiring across-task knowledge about \emph{how} to learn, e.g. by learning parameter initializations, learning rate schedulers, optimizers, etc. Usually there are two models involved: \emph{a base model} and \emph{a meta model}. A base model learns task-specific information from a small labeled training set (\emph{support set}) to predict on an unlabeled validation set (\emph{target set}). A meta model learns task-agnostic information to produce parameters for a base model enabling the fastest possible fine-tuning for each task at hand.

In this work we reproduce the paper of \cite{antoniou2019learning}, which proposes a framework, called \emph{Self-Critique and Adapt or SCA}, inspired by the idea that a target set also has task-specific information. For instance, if the training task is to distinguish cats from dogs and the new task would be to classify different breeds of dogs, a human would be able to guess what the new task is by observing a small number of samples. SCA aims at improving this ability by learning a label-free loss function during training, in order to be able to continue training the base model on the (unlabeled) test set before the final
inference.

The paper has been reproduced by solely reading the details therein, and not taking
the published code of the authors into account. It has become evident though, that
the published code is essential for a complete understanding of the work.

%The particular focus of this reproducibility study has been on the main contribution
%of the original paper, which is considered to be the use of SCA on top of existing
%state-of-the-art inner-loop optimization methods (e.g. MAML++).

\section{Background}
\subsection{Meta learning}
\begin{figure}[H]
    \centering
	\includegraphics[width=0.3\textwidth]{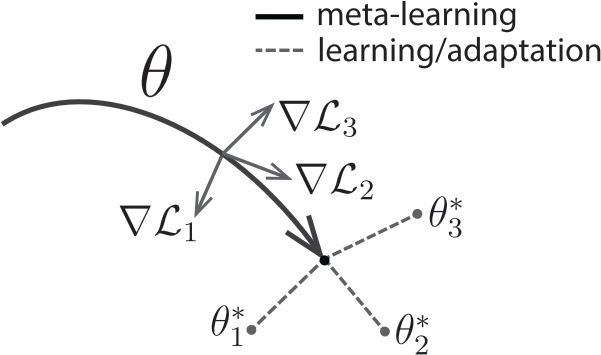}
    \caption{MAML diagram showing how base model parameters $\theta$ are optimized to quickly adapt to new tasks (taken from the original paper \citep{finn2017model})}
    \label{fig:maml}
\end{figure}

Meta learning has recently gained momentum after the publication of \emph{Model Agnostic Meta Learning} (MAML) \citep{finn2017model}. In short, MAML tries to optimize the initialization parameters $\theta$ of the base model, such that a network performs well on new few-shot learning tasks after only a few steps of training (see Figure \ref{fig:maml}). This approach, and its successors all have in common that a meta model is updated in an outer loop, while in an inner loop the task-specific base model is learned based on the meta model.
\vspace{-1em}
Versions of MAML include a first-order gradient version, denoted Reptile \citep{nichol2018reptile}, and a range of improvements presented by \cite{antoniou2018how}, denoted MAML++. In the latter, improvements were mainly focused at gradient instability, speedup of the first part of training by not using second-order gradients, learned step sizes, and tweaking Batch Normalization \citep{ioffe2015batch} for the meta-learning regime.

In short, given a task $b$, MAML++ updates the base model parameter vector $\theta_0^b =\theta_0$ by:
 \begin{equation}
 	\theta_{i+1}^b = \theta_i^b - \alpha \nabla_\theta\mathcal L_b(\theta_i^b)
 \end{equation}The loss for task $b$ is given by $\mathcal L_b$. The parameter $\theta_0$ is the parameter vector of the meta model, and is the common starting point, initialization, for all task-specific updates, as shown above. The meta model parameter vector is also updated, but in an outer loop, to minimize:
 \begin{equation}
 	\mathcal L_{\mathrm{meta}}(\theta_0) = \sum_{b=1}^B\sum_{i=1}^{N}v_i\mathcal L_b(\theta_i^b)
 \end{equation}
 Here, $B$ is the total number of tasks, and $N$ is the total number of base model updates. The scalars $v_i$ can be seen as importance weights which fulfill $v_i > 0$ and $\sum_i^{N} v_i = 1$. This is done to get more stable gradients by not only considering the last update, but throughout the entire update sequence. As training progresses, however, $v_N \rightarrow 1$ since the last update is ultimately what we care about and use during evaluation.
 
 Other interpretations of this algorithm, rather than meta learning, could be transfer learning since the common features of all tasks are encoded in the meta model, invariant of the task-specific updates. Yet another interpretation is that we learn a parameter initialization rather than randomly initializing network weights from some simpler distribution.

\input{sca_algorithm.tex}

\section{High-End MAML++}
\begin{figure}[H]
    \centering
	\includegraphics[width=0.9\textwidth]{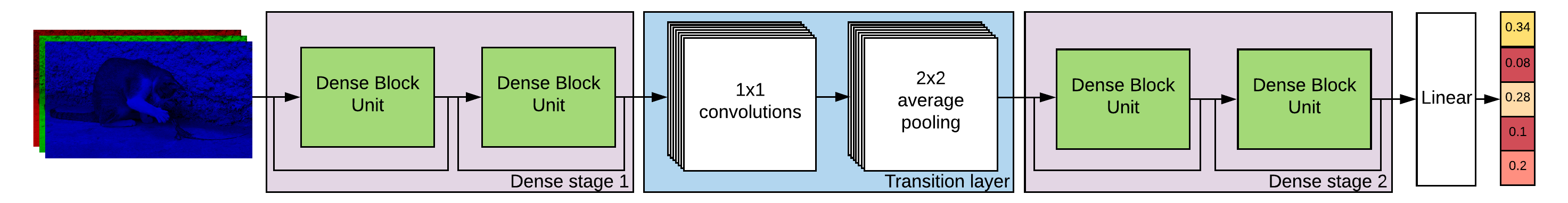}
    \caption{Architecture of the classifier in the High-End MAML++}
    \label{fig:he_maml}
\end{figure}

\begin{figure}[H]
    \centering
	\includegraphics[width=0.9\textwidth]{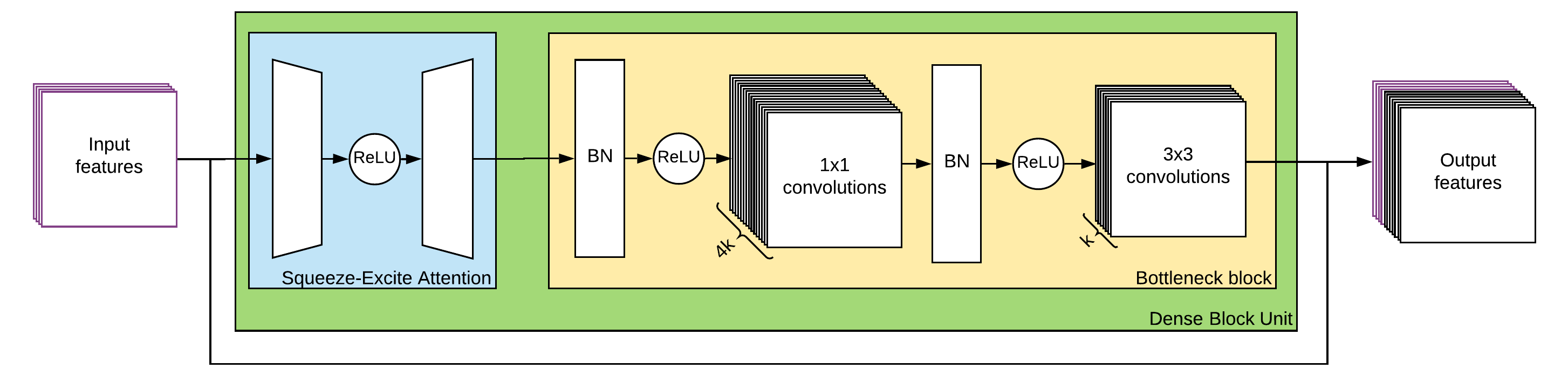}
    \caption{Architecture of the Dense block unit - a part of the classifier in the High-End MAML++. $k=64$ is a growth rate; BN stands for Batch Normalization.}
    \label{fig:he_dbu}
\end{figure}

In order to test whether SCA provides significant performance improvement for high-capacity models, the authors have introduced a novel high generalization performance MAML++ backbone, which is dubbed as \emph{High-End MAML++}. The main difference compared to the Low-End MAML++ is the use of a High-End Classifier (see architecture in Figure \ref{fig:he_maml}), instead of VGG network as the base model. The High-End classifier uses a DenseNet style architecture with 2 dense stages (purple blocks in Figure \ref{fig:he_maml}) and one transition layer in between them (light blue block in Figure \ref{fig:he_maml}). Each dense stage consists of two \emph{dense block units} (see architecture of dense block unit in Figure \ref{fig:he_dbu}). Each dense block unit consists of a bottleneck block, as described in \citep{huang2017densely}, preceded by the squeeze-excite style convolutional attention, as described in \citep{hu2018squeeze}.

\vspace{-1.5em}

To improve the performance and the training speed of the High-End classifier, the authors propose to optimize only the last dense block unit and the final linear layer in the inner loop. All the other network components are shared across inner loop optimization steps by treating them as feature embedding. Hence, all other components, \textbf{but the last dense block unit and the final linear layer}, will be optimized in the outer loop.

\section{Reproducibility}

\subsection{Critic architecture}
The key concept of DenseNets is that features of all previous layers are concatenated as inputs to all successive layers. For this to work, we either need the features to have the same size, or to preprocess them in some way such that the sizes agree. We decided to assume the sizes were kept constant by adding zero-padding. This padding will not be constant, however, since dilation changes per layer. This requires us to solve the equation
\begin{align}
    L &= \bigg\lfloor \frac{L + 2 \times \mathrm{padding} - \mathrm{dilation}(\mathrm{kernel\_size - 1) - 1}}{\mathrm{stride}} + 1 \bigg\rfloor \\
      &= \lfloor L + 2 \times \mathrm{padding} - 2^i \rfloor \\
\end{align}
where it is stated by the authors that: $i$ is the number of the layer from $0$ to $4$,
kernel size is $2$, and stride is $1$. $L$ is the size of the input and output. For the
first layer, we have:
\begin{equation}
    L = \lfloor L + 2 \times \mathrm{padding} - 1 \rfloor\\
\end{equation}
which has $\mathrm{padding} = 0.5$ as the solution. We solved this by adding one additional zero last in the feature dimension. For the remaining layers, the equation have integer solutions which makes the implementation trivial.

For the last fully connected layers, two questions are raised:
\begin{enumerate}
    \item Are all previous features from earlier layers propagated to the fully connected layer, or only to
the last convolutional layer?
    \item What is the feature dimension size after the first fully connected layer?
\end{enumerate}
For (1), we noticed that if skip connections to the fully connected layer were not present, the gradient of the output w.r.t. the input was zero. This changed after adding the skip connections and was therefore used in the experiments. For (2), in our experiments, we assumed that the feature size was kept after the first fully connected layer, making the weight matrix of this layer a square matrix. Another question raised was which activation function was used. We decided to use ReLU throughout the entire critic Network.

\subsection{Additional input features to the Critic}
The critic's input consists of three features: base model predictions on the target set, base model parameters used to produce these predictions and task embeddings produced by a neural network $g$. While supplying base model predictions as an input was not problematic, the other two input features were problematic to implement, because of the reasons, described below.

The embedding function $g$ was only partly specified by highlighting that a relational network produced superior results compared to DenseNet-style network. However, neither the architecture of such relational net nor its training procedure (i.e. which loss function was used, whether it was trained in conjunction with the meta and base model as in \citep{rusu2019meta}, etc) were explained. Due to the large amount of possible procedures for learning task embeddings, e.g., \citep{rusu2019meta,hausman2018learning,arnekvist2019vpe}, this part was not implemented.

Regarding passing base model parameters to the critic, the amount of required RAM becomes a problem. As an example, the base model in the case of Mini-ImageNet has roughly $70000$ parameters. In this case, the weights of the first fully connected layer of the critic would use
\begin{equation}
    (70000 \cdot (5 \cdot 8 + 1))^2 \cdot 32 \mathrm{bits} \approx 32 \mathrm{TB}
\end{equation}
of GPU memory, which is obviously not reasonable. In our experiments for adding base model parameters as features, we tried different sizes of the last fully connected layers, but all ran out of memory on a GPU with $11$GB RAM. Even when restricting the output dimension of the first fully connected layer to $256$, it still sums up to about $3$GB. Note, that this does not include the other parameters of the critic, the base/meta model, multiple forward activations, gradients, etc.

\subsection{Hyperparameters for Low-End MAML++}
In Algorithm \ref{algorithm_box}, there are three step sizes listed: $\alpha$, $\beta$, and $\gamma$.
We used the MAML++ approach where all the upgrades to the base model, including those made by the
critic, are done with vanilla SGD and learned step sizes. The critic's parameters were updated with SGD
and a relatively small step size $10^{-6}$. The reasoning was to use a relatively small step size and get slow learning, rather than using a large step size and risk unstable learning.

Batch sizes were $2$ for the $1$-shot experiments, as in MAML++, but $1$ for $5$-shot, since otherwise we
ran out of memory. In MAML++, it is proposed that Batch Normalization only normalizes data using the exponentional
moving averages, which implies that we can use batch size one.
As can be seen in section \ref{sec:results}, despite the changed batch size, results for Low-End MAML++
still turn out the same, or better than those reported in the original paper.

For the critic, we used standard initializations of layers defined in PyTorch
\citep{paszke2017automatic}, i.e, $w \sim \mathcal{U}(-\sqrt{k}, \sqrt{k})$
for the biases and weights of the convolutional layers, where $k^{-1}$ is the
number channels in times the kernel size. For the fully connected layers, we
used the same initialization but where $k^{-1}$ is the number of input features (fan-in).

\subsection{SCA for Low-End MAML++}
In SCA, we perform $I$ additional updates using the critic, after the $N$ inner loop updates using a
pre-defined loss function, usually negative log-likelihood. The simplest guess how to use MAML++ along with
SCA is by optimizing
 \begin{equation}
 	\mathcal L_{\mathrm{meta}}(\theta_0) = \sum_{b=1}^B\left[\sum_{i=1}^{N}v_i\mathcal L_b(\theta_i^b) + C(f(x_T^b, \theta_{N+I}^b))\right].
 \end{equation}
This can be extended in multi-step loss fashion (similar to MAML++), as follows:
 \begin{equation}
    \label{eq:multistep}
 	\mathcal L_{\mathrm{meta}}(\theta_0) = \sum_{b=1}^B\left[\sum_{i=1}^{N}v_i\mathcal L_b(\theta_i^b) + \sum_{j=1}^I w_j C(f(x_T^b, \theta_{N+j}^b))\right],
 \end{equation}
where $w_j$ are importance weights as $v_i$.

For the exact implementation details, the only information about the value of $I$ was given as an example in the Section 4 of the original paper, where $I=1$. For this reason, we used this value in our experiments, and then also the choice of importance weights $w_j$ for the critic losses becomes obsolete.

\subsection{Hyperparameters for High-End MAML++}

For most of the implementation details, the reader is referred to other papers \citep{huang2017densely,hu2018squeeze} and we had to assume that all details are the same as in these papers. Learning rate and optimizer for the High-End classifier are not specified, so we assumed that they should be the same as for Low-End MAML++.

However, when we were unable to reproduce the reported results with the assumed hyperparameters, we had to experiment further and tried different weight initialization strategies, namely Kaiming uniform and normal \citep{he2015delving}, Xavier \citep{glorot2010understanding}, as well as different optimizers, namely Adam optimizer \citep{kingma2014adam} and SGD. We noticed that these choices are crucial for the learning outcome, and after manual search we found SGD with learning rate $10^{-4}$ to be \emph{superior} to Adam. For weight initializations, we chose to follow the implementation of MAML++ with Xavier initialization and zero bias in all layers except the last linear layer of the classifier. If using Xavier in the last linear layer, interestingly, the learning immediately diverges. We instead found that the PyTorch \citep{paszke2017automatic} default initialization produced stable learning, but could not find any papers confirming this particular choice.

\section{Results}\label{sec:results}
We list our results in Table \ref{reprod-res}. For the original paper no results were reported for the Caltech-UCSD Birds 200 (CUBS-200) dataset \citep{WelinderEtal2010} using \emph{Low-End MAML++} (which is the original MAML++ proposed in \citep{antoniou2018how}), while we provide our results for this dataset here. The reported $95\%$ confidence intervals are calculated using the standard error times $1.96$. Standard deviations were estimated using the resulting accuracies of the $3$ experiments we ran, changing only the seed for each of them. Note, that in the original paper, neither a number of experimental runs, nor the way of calculating the confidence intervals were reported. 

The reported results for SCA are deemed \emph{significant} if the performance confidence intervals of MAML++ with SCA and without SCA do not overlap. Performance improvements that are considered significant are marked in bold in Table \ref{reprod-res}. Note, that the results are compared only \emph{within each implementation}, meaning that the reproduced SCA performance is compared to only the reproduced MAML++ performance and \emph{not} the one reported in the original paper.

It should be mentioned that we have also tried running SCA with High-End MAML++, but all experiments ran out of memory.

\begin{table}[H]
  \caption{SCA reproducibility results (all SCA models are applied on top of Low-End MAML++, the values in bold indicated significant improvements compared to the the base model)}
  \label{reprod-res}
  \centering
  \begin{tabular}{llllll}
    \toprule
    &&\multicolumn{4}{c}{\textbf{Test Accuracy}}\\
    \multicolumn{2}{l}{\textbf{Model}} & \multicolumn{2}{c}{\textbf{Mini-ImageNet}} & \multicolumn{2}{c}{\textbf{CUBS-200}} \\
    && \multicolumn{1}{c}{1-shot} & \multicolumn{1}{c}{5-shot} & \multicolumn{1}{c}{1-shot} & \multicolumn{1}{c}{5-shot} \\
    \midrule
    \multirow{2}{*}{\begin{minipage}{0.65in}MAML++\\(Low-End)\end{minipage}} & Orig. & $52.15 \pm 0.26\%$  & $68.32 \pm 0.44\%$ & - & -  \\
     & Ours & $51.41 \pm 0.20\%$ & $69.25 \pm 0.25\%$ & $55.62\pm1.57\%$ & $68.19\pm 0.86\%$  \\
    \midrule
    \multirow{2}{*}{\begin{minipage}{0.65in}with\\SCA (pred)\end{minipage}} & Orig. & $52.52 \pm 1.13\%$  & $\mathbf{70.84 \pm 0.34}\%$ &  - & -  \\
     & Ours & $\mathbf{55.38 \pm 0.39\%}$ & $70.42 \pm 1.25\%$ & $57.92\pm 1.55\%$ &  $\mathbf{69.96\pm 0.26}\%$ \\
    \midrule
    \multirow{2}{*}{\begin{minipage}{0.65in}MAML++\\(High-End)\end{minipage}} & Orig. & $58.37\pm 0.27\%$  & $75.50\pm 0.19\%$ &  $67.48\pm 1.44 \%$ & $83.80\pm 0.35 \%$  \\
     & Ours & $39.35 \pm 1.33\%$ & Out of memory & $31.23\pm 0.66\%$ & Out of memory \\
    \bottomrule
  \end{tabular}
\end{table}

All experiments for Mini-ImageNet were run on NVIDIA GeForce RTX 2080 Ti, all experiments for CUBS-200 were run on NVIDIA GeForce GTX 1080 Ti (both GPUs have 11 GB RAM). The running times for all experiments are reported in Table \ref{reprod-times}.
Note that we stop training after 10 epochs of no improvement in terms of accuracy on the validation set.

\begin{table}[H]
  \caption{Running time for the conducted reproducibility experiments}
  \label{reprod-times}
  \centering
  \begin{tabular}{lllll}
    \toprule
    \textbf{Model} & \multicolumn{2}{c}{\textbf{Mini-ImageNet}} & \multicolumn{2}{c}{\textbf{CUBS-200}} \\
    & \multicolumn{1}{c}{1-shot} & \multicolumn{1}{c}{5-shot} & \multicolumn{1}{c}{1-shot} & \multicolumn{1}{c}{5-shot} \\
    \midrule
    MAML++ (Low-End) & $2.6 \pm 0.8h$  & $2.8 \pm 0.1h$ & $1.1 \pm 0.2 h$ & $1.8 \pm 0.1 h $  \\
    \midrule
    \begin{minipage}{1.25in}MAML++ (Low-End)\\with SCA (pred)\end{minipage} & $7.8 \pm 0.6h$  & $5.2 \pm 0.7h$ &  $4.5 \pm 0.8 h$ & $3.7 \pm 0.3 h $  \\
    \midrule
    \begin{minipage}{1.25in}MAML++ (High-End)\end{minipage} & $6.0 \pm 0.8h$  & Out of memory &  $4.9 \pm 1.5 h$ & Out of memory  \\
    \bottomrule
  \end{tabular}
\end{table}

The source code for this re-implementation is built atop the MAML++ code published by \cite{antoniou2018how}, and can be found here: \url{https://github.com/dkalpakchi/ReproducingSCAPytorch}.

\section{Discussion}

The paper by \cite{antoniou2019learning} presents ideas that work well in practice for the Low-End MAML++, even without specific details that might be crucial to the success of other methods. The authors state that improvements to low-end methods often fail to give the same improvements to high-end
versions, and that they want to show that this is not true for SCA. It is mentioned that
meta-learning methods are very sensitive to architecture changes, nonetheless these details are explained in the paper \emph{very briefly}, totally excluding hyperparameter and parameter initialization details.
In fact, the paper is leaving out many details (which a reader will realize first in the middle of implementation).
It should be said though, that the authors have indeed published their code, which makes it an additional and
critical source of information. However, we have \textbf{not} consulted the published code, according to the guidelines of the NeurIPS Reproducibility Challenge.

In contrast to the reported results, we were not able to reproduce the results for the High-End MAML++, which can, of course, be caused by programming mistakes on our side. We did, however, employ pair programming which is a proven method to reduce errors \citep{hannay2009effectiveness} and spent ample time going through the code to spot bugs.

\vspace{16pt}

\subsection{Large amount of missing information}

We had to make several guesses on the architecture and hyperparameters, such as learning rates, weight initializations, and choice of optimizers. Although left out of the paper, we could mostly make reasonable guesses on how to implement it. On the other hand, for the embedding function $g$ too many details were absent from the paper to make it possible to implement. The initial part of the network was described briefly, but for the rest only a reference to a ``similar'' approach was mentioned \citep{rusu2019meta}. Among the details omitted, we feel that the most vital is the loss function for training the embedding function, especially since loss functions for embeddings are not as straightforward to guess as for regression and classification.

For the High-End backbone of MAML++, we noticed that different strategies for
weight initializations were crucial to be able to learn at all. In particular, following the
strategies in the MAML++ implementation and using Xavier initialization \citep{glorot2010understanding}
on the last linear layer before
the softmax led to a monotonically increasing loss. Instead using the standard initialization
of PyTorch \citep{paszke2017automatic} produced decreasing loss and increasing accuracy.

The reported results are based on versions of MAML++, but the only algorithm box in the paper describes SCA for MAML. Especially in conjunction with the missing number of critic updates $I$, it is hard to understand how to extend MAML++ to incorporate SCA, which gives rise to a bunch of questions. For instance, should the losses from the critic also be summed as for the $N$ previous steps? Should we use derivative order annealing as described in the MAML++ paper? Should we use importance weights if we go with a summation of the critic losses?

\vspace{16pt}

\subsection{Missing computational requirements}
The computational requirements for training any method play a vital role in the ability to use the method in practice. Unfortunately, such details are missing from the original paper making it hard to both estimate the computational power needed to reproduce the experiments and understand if the re-implemented model performs similarly to the original one.

\vspace{16pt}

\subsection{On the matter of ``magic'' numbers}
Some methods reported in the scientific literature might rely heavily on hyperparameters assuming
just the correct values, without reporting why these were chosen, how they were chosen, or in the
worst case not reported at all. When methods rely so heavily on particular hyperparameter choices,
one could either argue that they overfitted to the problem at hand and are not expected to
generalize beyond standard benchmarks, or that the method just happened to perform better because
it was allowed ample hyperparameter search. This raises a question of whether an alternative
method would be still better when allowed the same opportunity. For the reproduced paper though, we can
not draw any conclusions about whether crucial implementation details (e.g. details about architecture or other hyper-parameters) are missing or there are simply errors in our code.

In this work we have re-implemented SCA, for the Low-End MAML++ backbone, without specific
details (sometimes simply guessing), but we were still able to get results in favor of the method.
The fact that SCA worked even without the exact implementation details is a clear indication of good quality research. We were, however, not able to reproduce the results for the High-End
MAML++ and the experiments of High-End MAML++ with SCA ran out of memory. For the architectural details of the High-End backbone, the reader is mostly referred to read additional papers \citep{huang2017densely,hu2018squeeze} and we had no choice but to assume the exact same design choices as reported in the cited papers.
\newpage

\bibliography{references}
\end{document}

%% file: sca_algorithm.tex
\subsection{The SCA algorithm}
The paper we reproduce proposes a framework called Self-Critique and Adapt or SCA. The framework aims at using the information from the target set at inference time. This is achieved by learning a label-free loss function, parameterized as a neural network, called \emph{Critic Network} (later referred to as Critic). It is said that SCA can be applied for any meta-learning method that uses inner-loop optimization, but a specific example is given for the case of MAML (see Algorithm 1).

\SetKw{KwRequired}{Required:}
\begin{algorithm}[H]
\SetAlgoLined
\KwRequired{Base model function $\mathbf{f}$ and initialisation parameters $\bm{\theta}$, critic network function $\mathbf{C}$ and parameters $\mathbf{W}$, a batch of tasks $\{\mathbf{S}^{\mathbf{B}} = \{x_S^B, y_S^B\}, \mathbf{T}^{\mathbf{B}} = \{x_T^B, y_T^B\}\}$ (where $\mathbf{B}$ is the number of tasks in our batch) and learning rates $\alpha$, $\beta$, $\gamma$}
 
 \For{b in range($\mathbf{B}$)}{
    $\theta_0^b = \theta$\tcc*[r]{Reset $\theta_0$ to the learned initialization parameters}
    
    \For{i in range(N)\tcc*[r]{N is a number of inner loop steps wrt support set}}{
        \tcc{Inner loop optimization wrt support set}
        \hfill\llap{%
            \makebox[\linewidth]{\qquad\hspace{2.5em}$\theta_{i+1}^b = \theta_i^b - \alpha\nabla_{\theta_i^b}L(f(x_S^b, \theta_i^b), y_S^b)$\hfill\refstepcounter{equation}\llap{(\theequation)}\label{eq:sca1}}}
    }
    
    \For{j in range(I)\tcc*[r]{I is a number of inner loop steps wrt target set}}{
        \tcc{Critic feature-set collection}
        \hfill\llap{%
            \makebox[\linewidth]{\qquad\hspace{2.5em}$F = \{f(x_T^b, \theta_{N+j}^b), \theta_{N+j}^b, g(x_S^b, x_n)\}$\hfill\refstepcounter{equation}\llap{(\theequation)}\label{eq:sca2}}}
        
        \tcc{Inner loop optimization wrt target set}
        \hfill\llap{%
            \makebox[\linewidth]{\qquad\hspace{2.5em}$\theta_{N+j+1}^b = \theta_{N+j}^b - \gamma\nabla_{\theta_{N+j}^b}C(F,W)$\hfill\refstepcounter{equation}\llap{(\theequation)}\label{eq:sca3}}}
    }
    
    $L_{outer} = L_{outer} + L(f(x_T^b, \theta_{N + I}^b), y_T^b)$
 }
 \tcc{Joint outer loop optimization of $\theta$}
 \hfill\llap{%
    \makebox[\linewidth]{\qquad$\theta = \theta - \beta\nabla_\theta L_{\mathrm{outer}}$\hfill\refstepcounter{equation}\llap{(\theequation)}\label{eq:sca4}}}
 
 \tcc{Joint outer loop optimization of $W$}
 \hfill\llap{%
    \makebox[\linewidth]{\qquad$W = W - \beta\nabla_{W}L_{\mathrm{outer}}$\hfill\refstepcounter{equation}\llap{(\theequation)}\label{eq:sca5}}}
 
 \caption{SCA Algorithm combined with MAML}
 \label{algorithm_box}
\end{algorithm}

The major difference between SCA and MAML is the introduction of another inner loop part (lines 8 to 12 of Algorithm 1), where the base model weights are optimized with respect to the target set using the Critic Network $C$ as a loss function. The Critic operates on the collected feature set $F$ (see Equation \eqref{eq:sca2} of Algorithm 1), consisting of:
\begin{itemize}
    \item $f(x_T^b, \theta_{N+j}^b)$ - predictions of the base model (trained wrt support set) on the target set $T^b$ for the task $b$ using the base model parameters $\theta_{N+j}^b$;
    \item $\theta_{N+j}^b$ - parameters of the base model for the task $b$ after $N+j$ optimization steps;
    \item $g(x_S^b, x_n)$ - a task embedding, parameterized as a neural network, where $x_S^b$ is the support set for the task $b$ and $x_n$ is not described in the original paper.
\end{itemize}

These features serve as the input to the Critic $C$, which outputs the label-free loss function, used in the inner loop to perform gradient descent on the parameters $\theta$ of the base model (see Equation \eqref{eq:sca3} of Algorithm 1, where $W$ are parameters of the Critic and $\gamma$ is the Critic's learning rate). The loss for the outer loop $L_{\mathrm{outer}}$, used in original MAML now uses the base model parameters $\theta_{N+I}$ after $I$ updates using the Critic. The base model parameters $\theta$ (which will be used as initialization for the next task) and Critic's parameters $W$ are updated in the outer loop using $L_{\mathrm{outer}}$ loss. There are two key observations to make about the outer loop:
\begin{itemize}
    \item the base model parameters $\theta$ can be updated using the gradient of $L_{\mathrm{outer}}$, since the parameters $\theta_{N+I}^b$ depend on $\theta$ used to initialize $\theta_0^b$ (line 3 of Algorithm 1);
    \item the Critic parameters $W$ can be updated using the gradient of $L_{\mathrm{outer}}$, since $\theta_{N+I}^b$ depends on $W$ through gradient descent update (line 11 of Algorithm 1).
\end{itemize}

\begin{figure}[H]
    \centering
	\includegraphics[width=0.8\textwidth]{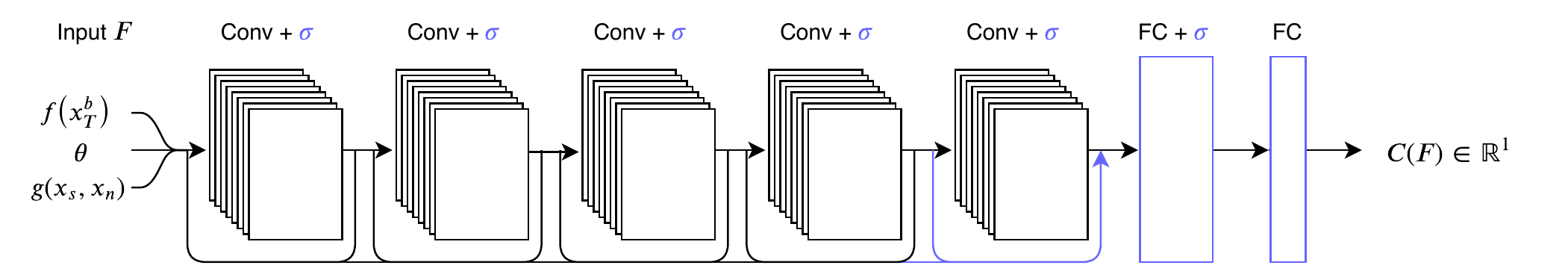}
    \caption{Critic Network architecture. The parts marked
    in blue correspond to parts which are not clear from
    the paper on how to implement. Other details that
    were unclear were padding or concatenation strategies
    and parameter initialization.}
    \label{fig:critic}
\end{figure}

\textbf{Critic Network Architecture.} The components of the Critic Network are shown in the Figure \ref{fig:critic}. As mentioned previously, the feature set $F$ serves as an input to the Critic. All features from $F$ are reshaped into a batch of 1D vectors, concatenated on the feature dimension and then passed to a sequence of five one-dimensional dilated convolutions with kernel size 2 and 8 kernels per layer. Each convolutional layer $i$ (starting from 0) uses an exponentially increasing dilation policy with a dilation $2^i$. Furthermore, DenseNet style connectivity is employed for convolutional layers, meaning that the input for each convolutional layer is a concatenation of outputs of all preceding convolutional layers. Finally a sequence of two fully-connected layers with ReLU activation functions is applied with the final fully-connected layer outputting a loss value. 
\vspace{-0.9em}
